\documentclass{article}

\usepackage{PRIMEarxiv}

\usepackage{enumitem} 
\usepackage[utf8]{inputenc} 
\usepackage[T1]{fontenc}    
\usepackage{hyperref}       
\usepackage{url}            
\usepackage{booktabs}       
\usepackage{amsfonts}       
\usepackage{nicefrac}       
\usepackage{microtype}      
\usepackage{lipsum}
\usepackage{fancyhdr}       
\usepackage{graphicx}       
\graphicspath{{media/}}     

\title{Unit: Building Unit Detection Dataset
}

\author{
  Haozhou Zhai,Yanzhe Gao,Tianjiang Hu\\
  School of Artificial Intelligence\\
  Sun Yat-sen University \\
  Zhuhai\\
  \texttt{\{zhaihzh, Gaoyzh25\}@mail2.sysu.edu.cn} \\
  \texttt{hutj3@mail.sysu.edu.cn} 
}

\begin{document}
\maketitle

\begin{abstract}
Fire scene datasets are crucial for training robust computer vision models, particularly in tasks such as fire early warning and emergency rescue operations. However, among the currently available fire-related data, there is a significant shortage of annotated data specifically targeting building units.To tackle this issue, we introduce an annotated dataset of building units captured by drones, which incorporates multiple enhancement techniques. We construct backgrounds using real multi-story scenes, combine motion blur and brightness adjustment to enhance the authenticity of the captured images, simulate drone shooting conditions under various circumstances, and employ large models to generate fire effects at different locations.The synthetic dataset generated by this method encompasses a wide range of building scenarios, with a total of 1,978 images. This dataset can effectively improve the generalization ability of fire unit detection, providing multi-scenario and scalable data while reducing the risks and costs associated with collecting real fire data. The dataset is available at https://github.com/boilermakerr/FireUnitData.
\end{abstract}


\section{Introduction}
\subsection{The Hazards of Fires and the Limitations of Tradition Detection Methods}
Fires are among the most devastating disasters worldwide. Fires cause substantial economic losses and casualties globally each year\cite{gaur2020video,geetha2021machine}.Traditional fire detection mainly relies on smoke sensors and temperature detectors, yet these two methods exhibit significant limitations. Firstly, sensors and detectors have a relatively slow response speed. They can only trigger alarms when the fire has developed to a certain extent, which may result in missing the optimal time for early warning and rescue operations.Secondly, sensors and detectors generally have a limited coverage area, making it difficult to achieve real-time monitoring of large areas as a whole.Additionally, since these two detection methods can only provide a single type of data (smoke concentration or temperature), they fail to offer intuitive information about the fire scene. As a consequence, emergency rescue efforts have to be carried out with insufficient information.

\subsection{Advantages of Unmanned Aerial Vehicles in Fire Detection}
In recent years, Unmanned Aerial Vehicles (UAV) technology has demonstrated distinct advantages in the field of fire monitoring.Firstly, UAVs are highly maneuverable and can be rapidly deployed in complex environments\cite{chen2022wildland}. They are particularly well-suited for scenarios that are difficult to cover with traditional equipment, such as high-rise buildings, factories, and mountainous areas. Secondly, modern UAVs are equipped with high-definition cameras and infrared sensors, which can provide multi-dimensional fire information. Visible light images can be used to display the location and spread of the fire source, while thermal imaging data helps identify high-temperature areas. More importantly, UAV systems can integrate computer vision algorithms to achieve intelligent fire detection. Through deep learning models, the application of UAVs in fire-related tasks has advanced beyond mere "detection" to "analysis". The UAV system can not only identify flames but also assess the severity level of the fire, providing data support for rescue decision-making.

\subsection{The Importance of Unit-Level Localization} 
Achieving unit-level localization to accurately determine the specific unit where a fire occurs is crucial for enhancing emergency response efficiency. Traditional detection methods typically can only determine whether a fire has broken out in a building, whereas modern intelligent systems can provide more precise spatial information. Their advantages are mainly manifested in three aspects: Firstly, in terms of improving rescue efficiency, firefighters can obtain prior knowledge of the fire's origin and plan the optimal rescue route in advance. For example, in a high-rise residential building fire, a unit-level localization system can assist rescue personnel in bypassing dangerous areas such as elevator shafts and directly reaching the unit where the fire has started, thereby saving rescue time. Secondly, with regard to the optimal allocation of resources, by linking with smart city systems, the nearest firefighting equipment and emergency resources can be automatically dispatched. This enables a fire protection system integrated with unit-level localization to shorten the emergency response time. Thirdly, concerning the optimization of evacuation guidance, accurate fire location information helps in formulating more targeted evacuation plans, avoiding the uncertain risks associated with blind evacuation.

\subsection{Current Status of Fire Datasets}
At present, mainstream fire datasets\cite{wu2023dataset,
shamsoshoara2021aerial,
hopkins2023flame,
wang2025open,
chen2022wildland} demonstrate outstanding performance and yield notable results in macro-level aspects such as overall fire scene recognition, fire trend prediction, and fire risk assessment. Supported by large-scale data samples and coverage of diverse fire scenarios, these datasets provide a solid data foundation for fire science research. However, there is a significant lack of unit - level information in existing mainstream fire datasets. This restricts the further improvement in the application of fire detection algorithms, especially in complex building environments. Therefore, we propose a fire detection dataset targeting building units.

\subsection{Contributions of This Study}

In response to the aforementioned issues, we propose a method for constructing a fire dataset. The main contributions are as follows:
\begin{itemize}[leftmargin=*] 
\item Multi-scenario data collection: We employed DJI drones to collect data from six different scenarios, resulting in a dataset that encompasses eight types of buildings.
\item Building facade region extraction: We extracted the facade regions of buildings, eliminating interference from other buildings and the background, thereby focusing the data on the targets to be detected.
\item Dynamic enhancement processing: We simulated various effects that drones may encounter during flight, including illumination enhancement, motion blur, and the generation of flames in specific regions using large image models.
\end{itemize} 
The final constructed dataset contains 1,978 sets of high-quality samples, with each set comprising an original image and an annotation file in TXT format.

 \begin{figure*}[!ht]
\vspace{3mm}

\centering

\centerline{\includegraphics[width=\textwidth,height=8cm,keepaspectratio]{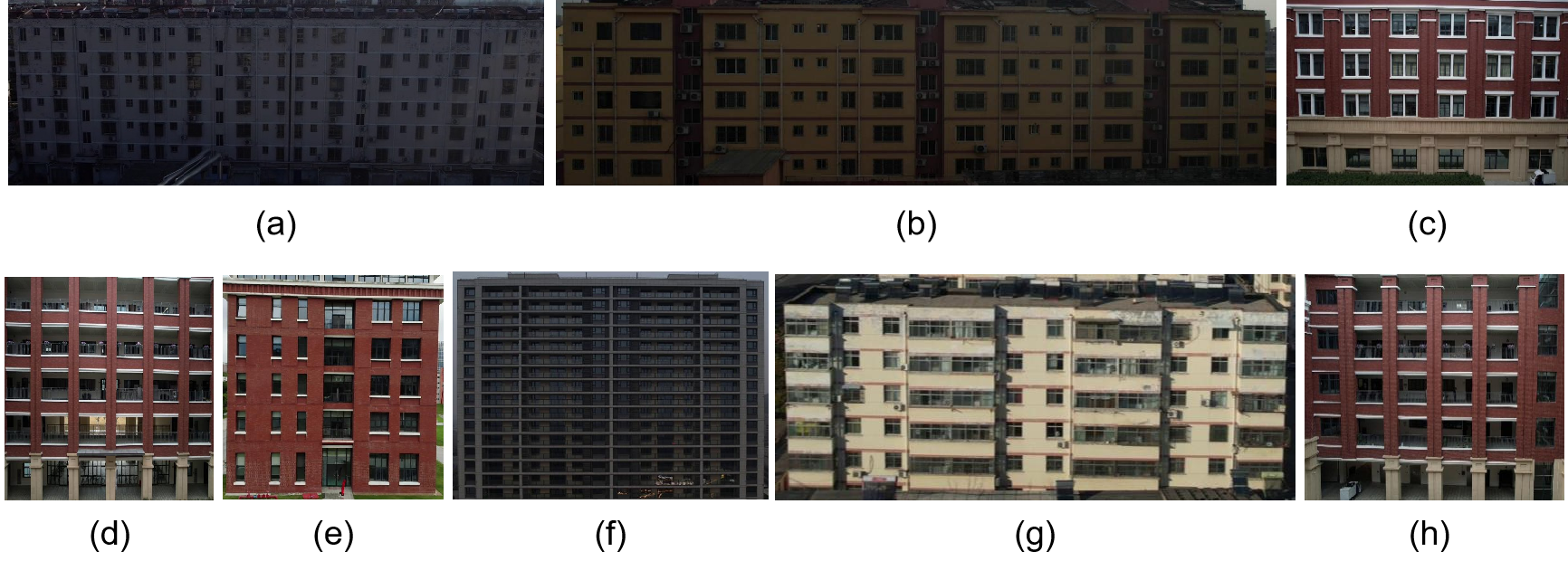}}
\caption{Overview of diverse building in the dataset}
\label{building overview}
\end{figure*}
\section{Details of the Dataset }

The dataset contains a total of 1,978 data samples, with 1,777 in the training set and 201 in the test set. To ensure the diversity of the dataset, we collected high-definition images of eight different building facades from six regions. As shown in Fig. \ref{building overview}, the collected objects include high-rise buildings, multi-story buildings, row houses, and standalone buildings, each with the following characteristics:

High-rise buildings: These are typically defined as residential buildings with a height of 10 or more stories, as shown in Fig. \ref{building overview} (f). They feature significant building heights and a large number of vertical units. The units are generally densely packed and exhibit a high degree of similarity in their structural designs.

Multi-story buildings: These usually refer to buildings with fewer than 10 stories, such as those depicted in Fig. \ref{building overview} (c), (d), (e), and (h). They have a relatively small number of vertical units. Some buildings have distinct window structures on their facades, with variations in window types and the distribution of facade units.

Row houses: Row houses are composed of multiple building units connected horizontally in close proximity, as illustrated in Fig. \ref{building overview} (a), (b), and (g). They typically have a wide overall facade width and are commonly found in residential areas. The sizes of individual units vary, but they are arranged in a somewhat regular pattern.

Standalone buildings: These are independent structures with no direct physical connection to other buildings, as seen in Fig. \ref{building overview} (d) and (e). They generally have a relatively narrow overall facade width. Most of the building units have a consistent form, although a few units may exhibit significant differences in size.

\section{Data Augmentation Techniques}

 \begin{figure*}[!ht]
\vspace{3mm}

\centering

\centerline{\includegraphics[width=\textwidth,height=8cm,keepaspectratio]{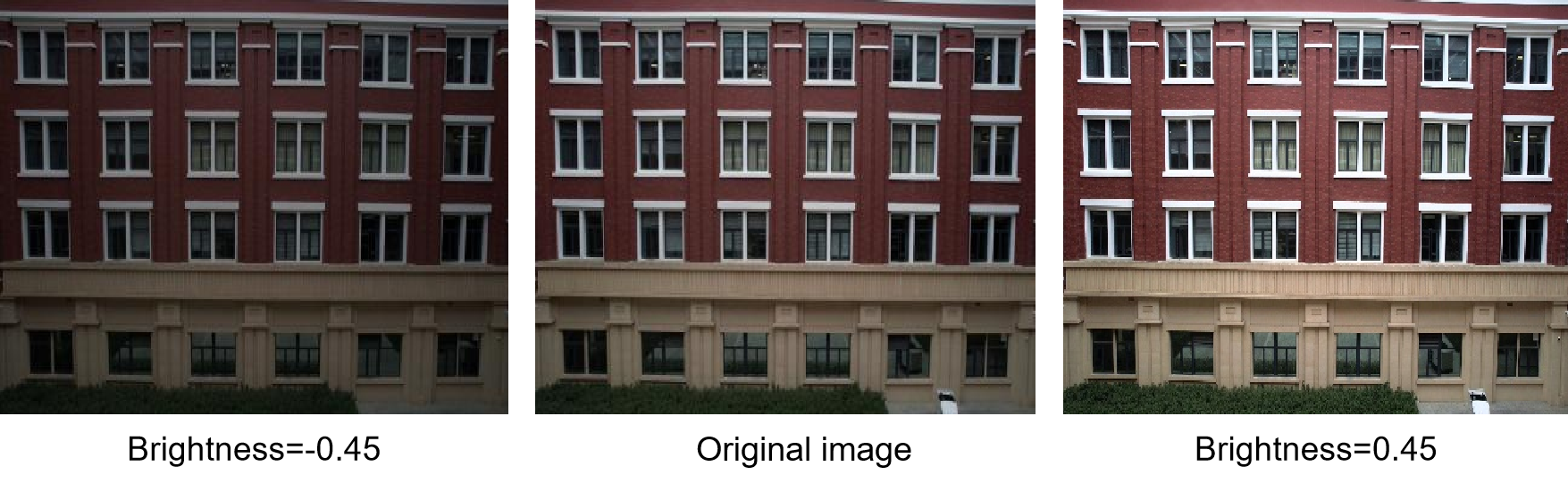}}
\caption{Illumination variations under different brightness values effect}
\label{light}
\end{figure*}

 \begin{figure*}[!ht]
\vspace{3mm}

\centering

\centerline{\includegraphics[width=\textwidth,height=6cm,keepaspectratio]{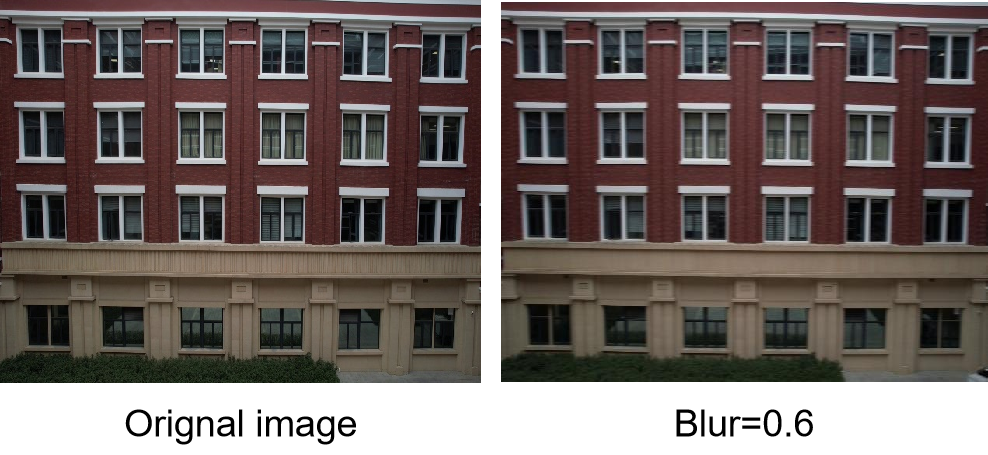}}
\caption{Motion blur effect under blur values}
\label{ai}
\end{figure*}

 \begin{figure*}[!ht]
\vspace{3mm}

\centering

\centerline{\includegraphics[width=\textwidth,height=6cm,keepaspectratio]{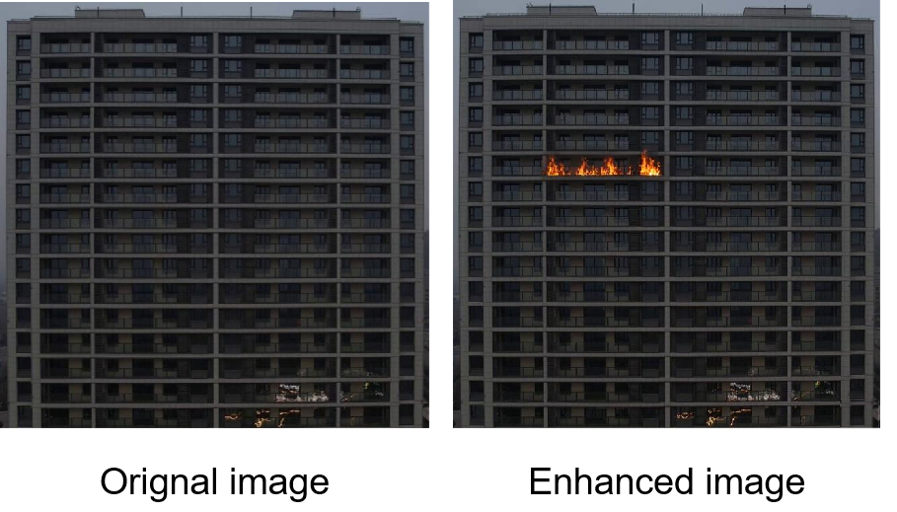}}
\caption{Orignal image and AI-generated fire in selected region}
\label{blur}
\end{figure*}

To simulate the differences in images captured under various flight conditions during UAV operations, we employed a motion blur synthesis algorithm to replicate the blurring effects that occur during UAV flight. This was achieved by applying a convolution operation using OpenCV's filter2D function, with the size of the blur kernel controlled by specific parameters. The weight distribution within the convolution kernel ensures that the value of each pixel is influenced by its surrounding pixels. By adjusting the kernel size and weight distribution, we can selectively suppress high-frequency components of the image while preserving low-frequency structural information, aligning with the frequency-domain attenuation characteristics of real motion blur. As shown in Fig. \ref{blur}, the algorithm supports parameterized adjustment of blur intensity, allowing it to accommodate imaging differences under various flight states, such as hovering, cruising, and maneuvering. This method not only ensures efficient generation but also produces physically plausible blurring effects, effectively enhancing the model's robustness against motion-degraded images. Additionally, by controlling the parameters, we can freely adjust the degree of blur, thereby covering a wider range of flight-induced imaging scenarios to a greater extent.

To enhance the generalization capability of the dataset and simulate illumination variations in real-world scenarios, we designed a data augmentation system for illumination transformation. This system generates a diverse set of images with varying brightness levels by applying photometric transformations to the original scenes. Specifically, we employ color space conversion technology, transforming the images from the BGR color space to the HSV (Hue-Saturation-Value) color space. Within the HSV space, we perform linear scaling operations on the luminance component (V channel) to achieve controllable brightness adjustments. After completing the photometric adjustments, the images are converted back to the BGR color space through inverse color space conversion, ultimately generating an augmented dataset with illumination diversity.  As shown in Fig. \ref{light}, This method allows for precise brightness control while maintaining constant hue and saturation, aligning with human visual perception characteristics and enabling the generation of continuous illumination variation samples. By utilizing photometric transformation approach, we effectively improved the model's robustness under different illumination conditions.

 As shown in Fig. \ref{ai}, to generate highly realistic flame-augmented data, we leveraged large-scale image generation models. First, irregular mask regions are randomly generated on the original images to serve as target locations for flame synthesis. Subsequently, the large-scale model is employed to generate flames within these mask regions, with the generation process encompassing flame characteristic simulation and natural integration with the building structures. This method can be extended to other disaster simulations, enhancing the model's generalization capability in complex scenarios.

\bibliographystyle{unsrt}  
\bibliography{main}

\end{document}